\PassOptionsToPackage{table}{xcolor}
\documentclass{article} %
\usepackage{iclr2026_conference,times}
\iclrfinalcopy

\usepackage[T1]{fontenc}

\usepackage{amsmath,amsfonts,bm}

\def\eqref#1{equation~\ref{#1}}

\def\1{\bm{1}}

\DeclareMathAlphabet{\mathsfit}{\encodingdefault}{\sfdefault}{m}{sl}
\SetMathAlphabet{\mathsfit}{bold}{\encodingdefault}{\sfdefault}{bx}{n}

\usepackage{natbib}
\usepackage{graphicx}
\usepackage{subcaption}
\usepackage{wrapfig}
\usepackage{needspace}
\usepackage{booktabs}
\usepackage{algorithm}
\usepackage{algpseudocode}
\usepackage{xcolor}
\usepackage{tabularx}
\usepackage{longtable}
\usepackage{float}
\usepackage{url}
\usepackage{hyperref}
\hypersetup{colorlinks=true,allcolors=black}

\newcolumntype{Y}{>{\centering\arraybackslash}X}

\title{Scaling Video Generation for Reasoning: \\At What Cost?}

\author{%
Weihang Guo$^{1}$ \quad
Xiaoyu Wu$^{2}$ \quad
Yifei Wang$^{1}$ \quad
Niloofar Mireshghallah$^{2}$ \quad
Lydia E. Kavraki$^{1}$ \\[0.4em]
$^{1}$Rice University \qquad
$^{2}$Carnegie Mellon University \\
\texttt{wg25@rice.edu, xiaoyuwu@andrew.cmu.edu,  yw251@rice.edu} \\
\texttt{nmireshg@andrew.cmu.edu, kavraki@rice.edu} \\
}

\begin{document}

\maketitle

\begin{abstract}
We study whether scaling video generation enables models to reason about
hidden information from the past frames, and at what computational cost. Our controlled benchmark
requires predicting nine prescribed moves of an initially solved
$2\times2\times2$ Rubik's Cube from a fixed view of three faces.
Correct predictions require inferring how actions change hidden states,
and the simulator provides exact ground truth for evaluation.
Models learn plausible cube geometry early,
while correct sticker configurations require substantially more training.
Although validation MSE follows approximate power-law scaling, lower MSE loss
does not reliably indicate downstream reasoning capabilities.
Smaller autoregressive models achieve higher state accuracy with
limited compute, while larger models reach higher accuracy after more
training. At roughly 0.1 PF-days, the 70M-parameter model correctly predicts
the visible sticker configuration in 44.6\% of post-action frames, compared
with 0.3\% for the 1B model, which reaches 83.7\% at 3.14 PF-days.
Symbolic state supervision raise the 20M
model's frame accuracy from 31.1\% to 67.3\% at the same training-data
budget, suggesting that learning representations of state changes can
complement scaling.

\begin{figure}[h]
    \centering
    \includegraphics[width=0.9\linewidth]{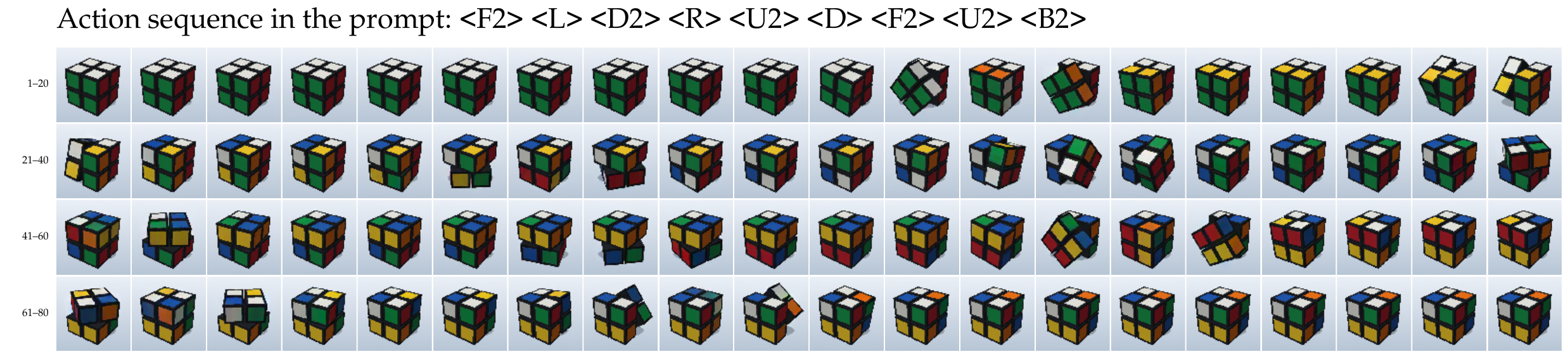}\par\nointerlineskip\vspace{1pt}
    \includegraphics[width=0.9\linewidth]{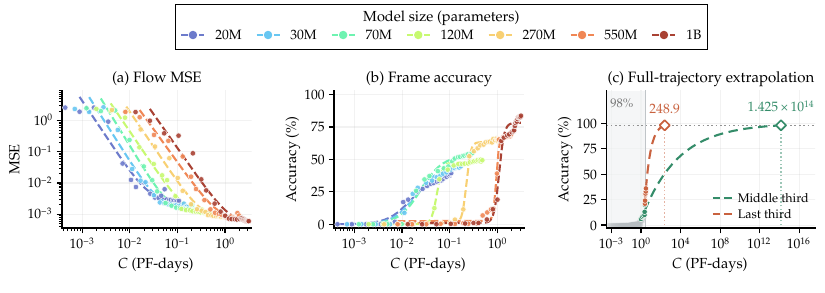}
    \caption{Rubik video generation and scaling with training compute.
    The video strip shows 80 predicted frames, or 3.3~s at 24~fps, \textbf{for 9
    prescribed actions as the language prompt from a solved cube}. The video generator is trained
    from scratch on synthetic videos. (a) Validation flow MSE and
    (b) average frame accuracy for AR models of different sizes.
    (c) AR full-trajectory accuracy, which requires all nine post-action
    states to be correct. Separate power-law fits to the middle and final
    thirds of its observed compute frontier illustrate how compute demand
    grows toward higher reliability. Diamonds mark their window-dependent
    98\% intersections.}
    \label{fig:rubik-scaling-overview}
\end{figure}
\end{abstract}

\section{Introduction}

Video generative models have attracted considerable attention for their
ability to produce realistic videos for content creation and
editing~\citep{wanteam2025wan,kong2024hunyuanvideo}.
Their ability to learn patterns of motion and interaction from video
has also motivated their use as predictive models of the physical
world~\citep{yang2024video,yang2024learning,ball2025genie3}.
This perspective extends their role from generating visual content to
anticipating how an environment may change, with applications in
robot control and action-conditioned world
models~\citep{ye2026world,kim2026cosmos,li2026lingbova,
wang2026interactive,nvidia2025cosmos}.

\begin{figure}[h]
    \centering
    \includegraphics[width=\linewidth]{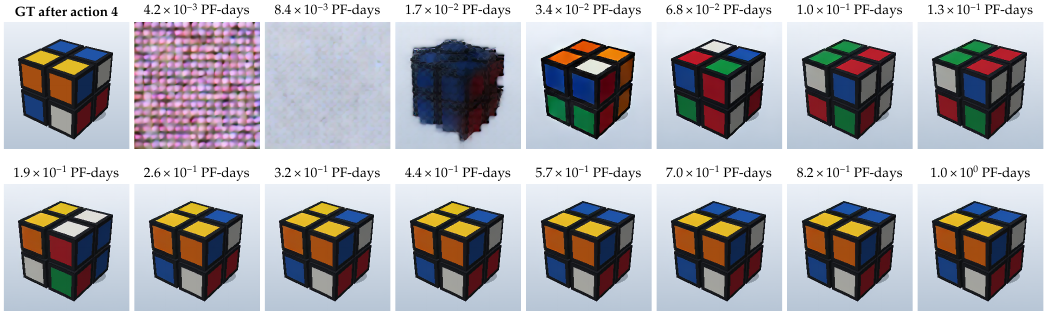}
    \caption{Predictions from the 270M-parameter autoregressive model after
    four prescribed moves from a \emph{solved cube}.
    The model reproduces the cube's geometry after
    $3.4\times10^{-2}$ PF-days of training, but first predicts all 12 visible
    stickers correctly in this example at $5.7\times10^{-1}$ PF-days,
    requiring about 17 times as much compute. The top-left image is the
    ground truth.}
    \label{fig:rubik-checkpoint-progression}
\end{figure}

A visually convincing video can still depict an outcome that contradicts
preceding events. For example, in
Figure~\ref{fig:rubik-checkpoint-progression}, early predictions already
reproduce the cube's geometry, but its sticker colors remain inconsistent
with the prescribed moves. Under partial observation, correct prediction
may require information from earlier frames and inference about how
intervening actions have changed hidden
states~\citep{kaelbling1998planning,reiter2001knowledge}.
Existing scaling studies relate prediction loss to model size, data, and
compute~\citep{kaplan2020scalinglaws,liang2026diffusionscaling,yin2025videoscaling}.
These loss trends alone do not establish whether state predictions become
more accurate. We therefore ask: \textbf{Does lower validation flow MSE
reliably indicate better reasoning about hidden states across models?
How does computational cost grow as state prediction becomes more reliable?}

To investigate these questions, we devise a controlled video prediction
task to study how models reason about hidden state changes. Each video
starts from a solved $2\times2\times2$ Rubik's Cube and follows nine
prescribed moves. A fixed camera shows three faces under consistent
rendering conditions. The moves rearrange both visible and hidden
stickers, so predicting the colors that reappear requires accounting for
the intervening actions.
This setup provides a direct way to evaluate state prediction: \textbf{the initial
frame and action sequence determine every future state}, allowing
generated observations to be checked against exact ground truth.
Sampling action sequences also provides an effectively unlimited supply
of training videos, allowing us to scale training with fresh data.
In the scaling experiments, each model is trained for a single epoch,
using each training video only once.

Using this task, we train bidirectional (Bidir) and autoregressive (AR)
video diffusion transformers from scratch at seven model sizes from 20M
to 1B parameters, varying the amount of training data. We measure
validation flow MSE alongside action
following and state accuracy in generated videos, using the metrics in
Table~\ref{tab:rubik-metric-definitions}. This comparison allows us to
examine whether improvements in the training objective translate into
more accurate state prediction, and how these outcomes scale with data
and compute (Figure~\ref{fig:rubik-scaling-overview}).
We also compare video prediction with direct prediction of discrete
sticker states and expose all six faces to examine the effect of partial
observation.

Our main findings are:
\begin{enumerate}
    \item Lower validation flow MSE does not reliably indicate higher
    action or state accuracy across model sizes. After 1.5M training
    videos, the 1B AR model has lower MSE than the 70M model, yet its
    average frame accuracy is only 3.67\%, compared with 41.00\%.
    \item Additional training gives diminishing returns at a fixed model
    size. For the 120M AR model, increasing training data from 5M to 8M
    videos raises average frame accuracy only from 48.11\% to 49.33\%.
    Larger models improve accuracy, while extrapolations of frame and
    full-trajectory error point to substantial further compute as
    prediction becomes more reliable.
    \item State supervision and predicted-state feedback improve video
    accuracy at the same training-data budget. For a 20M AR-$k=1$
    video model trained on 3M videos, state guidance raises average
    frame accuracy from 31.1\% to 67.3\%. This supports learning
    representations for state reasoning as a complement to scaling.
\end{enumerate}

\section{A Controlled Study of State Prediction}
\label{sec:controlled-study}

\subsection{State Prediction under Partial Observation}

\begin{figure}[h]
    \centering
    \includegraphics[width=\linewidth]{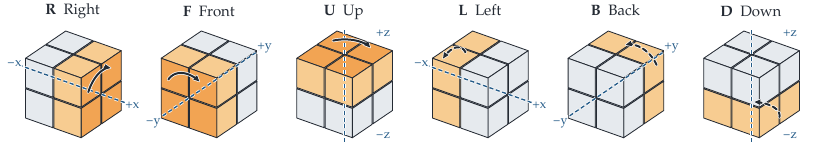}
    \caption{Rubik face-turn notation. Orange cubies rotate around the blue
    axes. Arrows show $90^\circ$ clockwise turns viewed from outside the named
    face. Dashed arrows mark hidden faces. A prime reverses the turn, and a
    suffix $2$ denotes $180^\circ$.}
    \label{fig:rubik-action-axes}
\end{figure}

We study a solved $2\times2\times2$ Rubik's Cube undergoing nine prompted
face turns, using the notation in Figure~\ref{fig:rubik-action-axes}. A fixed
camera shows three faces, exposing 12 of the cube's 24 stickers at each
settled state. Turns move stickers between visible and hidden faces. A sticker
can leave view and later return at a different position, so the colors that
should appear after a turn depend on the preceding action sequence. The task
is to predict these changing observations as the prescribed moves are
executed.

Our action set contains 18 moves: the six clockwise $90^\circ$ turns shown in
Figure~\ref{fig:rubik-action-axes}, their counterclockwise inverses, and the six $180^\circ$ turns. We
extend the ModernBERT tokenizer~\citep{warner2025modernbert} with one
dedicated token for each move.

\textbf{All videos always start with a solved configuration, and a fixed color convention specifies the entire
initial state, including the hidden faces.} Each face turn is a deterministic
permutation of stickers, so the initial state and action sequence uniquely
determine every subsequent state. Keeping the camera and rendering fixed
gives an unambiguous visual target for each transition. This allows us to
measure whether generated observations reflect the correct state changes,
while leaving the model free to learn its own internal representation.

\subsection{Action-Conditioned Video Prediction}

The video generator receives the initial image and a prompt containing the
complete ordered action sequence. A frozen Wan2.1 VAE~\citep{wanteam2025wan}
encodes the video into latent frames, and a frozen ModernBERT encoder
supplies language features. A 
DiT~\citep{peebles2023scalable} predicts future video latents conditioned
on the initial visual latents and language features. We train the DiT from
scratch using rectified flow matching~\citep{liu2023rectifiedflow}, minimizing
the MSE between predicted and target latent velocities. This combination of
latent video DiTs and flow-matching objectives is also used in
Wan~\citep{wanteam2025wan}, HunyuanVideo~\citep{kong2024hunyuanvideo},
LTX-Video~\citep{hacohen2024ltxvideo}, and
Cosmos-Predict2.5~\citep{nvidia2025cosmospredict25}.

We include bidirectional~(Bidir) and autoregressive~(AR) generation. The bidirectional
model generates all future latent frames jointly. The autoregressive model
generates chronological chunks of $k$ latent frames, conditioning each chunk
on the preceding visual history; we denote this variant by AR-$k$. Both
variants use three-dimensional rotary position embeddings
(RoPE)~\citep{su2021roformer} to encode the temporal and spatial coordinates
of visual tokens. During AR training, the history consists of ground-truth
latents. During generation, the model uses its own previous predictions,
allowing earlier errors to affect later parts of the rollout.
For both variants, we maintain an exponential moving average (EMA) of the
generator parameters during training and use the EMA weights for evaluation
and generation.
Appendix~\ref{app:video-dit-scaling} gives the architecture and training
settings; Algorithm~\ref{alg:generator-training} in
Appendix~\ref{app:training-procedure} summarizes the training procedure.

Figure~\ref{fig:rubik-checkpoint-progression} illustrates the distinction
between visual plausibility and state accuracy: recognizable cubes can still
have incorrect sticker configurations.

\subsection{Evaluating State Prediction}

We evaluate generated rollouts using the metrics defined in
Table~\ref{tab:rubik-metric-definitions}. At the first stationary frame after
each turn, we decode sticker colors from the image regions specified by the
simulator and compare them with the expected configuration. Unreadable
stickers count as incorrect. The sticker and frame scores distinguish
partial state recovery from a completely correct visible configuration.
VAE reconstruction preserves all scored sticker states in the 100
ground-truth evaluation videos (Appendix~\ref{app:benchmark-games},
Table~\ref{tab:rubik-vae-state}).

We measure action following separately with a
motion probe trained on grayscale simulator videos. The probe predicts the
moved face and turn type from optical flow, allowing us to distinguish
errors in action execution from errors in the resulting state. Evaluator
details appear in Appendix~\ref{app:benchmark-games}.
All accuracy metrics are measured on free-running videos, with no
ground-truth history supplied after the initial image. Validation flow MSE
is measured separately on held-out videos under the training conditions. It measures velocity
prediction in latent space, while the accuracy metrics assess the states
reached during generation.

\begin{table}[h]
    \centering
    \small
    \caption{Evaluation metrics.}
    \label{tab:rubik-metric-definitions}
    \begin{tabularx}{\linewidth}{@{}l>{\raggedright\arraybackslash}X@{}}
        \toprule
        \textbf{Metric} & \textbf{Definition} \\
        \midrule
        MSE loss &
        Mean squared error between the predicted and target flow velocities
        in video latent space on \emph{validation} set. \\
        Action following acc. &
        Fraction of actions in the prompt that are correctly applied to the
        cube. \\
        Sticker acc. &
        Fraction of \emph{visible} stickers ($4$ stickers $\times$ $3$ visible faces)
        with the correct color after each action. \\
        Frame acc. &
        Fraction of the nine post-action frames in which \emph{all} 12 visible
        stickers have the correct color. \\
        Full-trajectory acc. &
        Fraction of episodes in which all 12 visible stickers are correct
        after every one of the nine actions. \\
        \bottomrule
    \end{tabularx}
\end{table}

\section{The Scaling Experiments}
\subsection{Experimental Setup}
\paragraph{Dataset generation.}
We generate a shared stream of Rubik videos for the architecture and scaling
experiments. Each video contains nine actions over 81 frames at
$256\times256$ resolution. Actions are sampled uniformly from the legal
moves, excluding consecutive turns of the same face. The architecture
comparison uses the first 1M
videos. Each model in the scaling experiments has a budget of 8M videos.
All models consume the same ordered stream, using each training video once.
Architecture selection and scaling evaluation use the same 100 held-out
episodes, paired across models and checkpoints. Validation MSE curves use a fixed subset of 256
held-out videos, expanded to 1,024 for the architecture comparison endpoints.

\paragraph{Video DiT architecture.}
We first compare three depth--width configurations at approximately 95M
parameters (Table~\ref{tab:stage-a-architecture}). Each is trained with
autoregressive chunks of one or four latent
frames, or with bidirectional generation, giving nine configurations in total.
All runs use the same 1M training videos, optimizer settings, and effective
batch size.
Both the depth--width allocation and the generation scheme strongly affect
accuracy at this budget. Frame
accuracy improves as capacity shifts from depth to width, and one-frame
autoregressive generation performs best within each depth--width
configuration. The wide--shallow AR-$k=1$ model leads on action, sticker,
and frame accuracy. We therefore carry the
wide--shallow AR-$k=1$ and Bidir configuration into the scaling experiments. 
Further results and training details appear in Appendix~\ref{app:stage-a-results}.

\begin{table}[h]
    \centering
    \small
    \caption{Stage A architectures and results after 1M training videos.
    Action, sticker, and frame accuracies (\%) use the same 100 held-out videos.
    AR-$k$ generates $k$ latent frames per chunk. Best accuracies are bold.}
    \label{tab:stage-a-architecture}
    \label{tab:architecture-comparison}
    \setlength{\tabcolsep}{2.5pt}
    \begin{tabular}{@{}lrrrrrlrrr@{}}
        \toprule
        Configuration & $d_{\mathrm{model}}$ & Layers & Heads & $d_{\mathrm{ff}}$ & Parameters (M) & Generation & Action & Sticker & Frame \\
        \midrule
        Deep-narrow  & 512 & 22 & 8 & 1408 & 95.03 & AR-$k=1$ & 17.22 & 25.37 & 3.67 \\ %
                     &     &    &   &      &            & AR-$k=4$ & 6.33  & 22.20 & 0.78 \\
                     &     &    &   &      &            & Bidir    & 7.33  & 21.59 & 0.33 \\
        \midrule
        Balanced     & 640 & 14 & 10 & 1728 & 94.20 & AR-$k=1$ & 72.44 & 52.00 & 24.33 \\ %
                     &     &    &    &      &            & AR-$k=4$ & 9.67  & 22.46 & 1.67 \\
                     &     &    &    &      &            & Bidir    & 6.78  & 21.55 & 0.56 \\
        \midrule
        \textbf{Wide-shallow} & 768 & 10 & 12 & 2048 & 96.91 & AR-$k=1$ & \textbf{90.78} & \textbf{78.20} & \textbf{37.22} \\ %
                     &     &    &    &      &            & AR-$k=4$ & 79.89 & 63.14 & 25.11 \\
                     &     &    &    &      &            & Bidir    & \textbf{8.33}  & \textbf{21.67} & \textbf{0.78} \\
        \bottomrule
    \end{tabular}
\end{table}

\paragraph{Scaling up video DiT.}
We scale the selected architecture family across seven model sizes, from
approximately 20M to 1B trainable parameters, by increasing width and depth.
AR-$k=1$ and bidirectional models share the same architecture dimensions and
parameter counts at each size. Detailed specifications appear in
Appendix~\ref{app:video-dit-scaling}, Table~\ref{tab:scaling-runs}.
The frozen video and language encoders remain unchanged. All models
use the same training stream, effective batch size, and optimizer settings.
This allows us to compare additional training at a fixed model size and
larger models at a fixed data budget.
We evaluate checkpoints throughout training, comparing model sizes at a
common exposure of 8M videos. To compare these choices under a common resource
budget, we also estimate training compute in FLOPs, accounting for both
model size and data exposure.
Appendix~\ref{app:compute-training} gives the compute accounting.

\subsection{Understanding the Difficulty of Hidden State Prediction}
Before examining how performance scales with model size and training data, we first ask what makes the task difficult. In the following paragraphs, we show that much smaller models readily learn the cube's state transitions in a compact and discrete vector space. Then, we show that revealing the full cube state substantially improves video prediction. Together, these comparisons suggest that a central challenge lies in tracking and reasoning hidden states through video generation.

\paragraph{Small models learn state reasoning in discrete vector space}
\label{sec:state-only}

\begin{figure}[h]
    \centering
    \includegraphics[width=\linewidth]{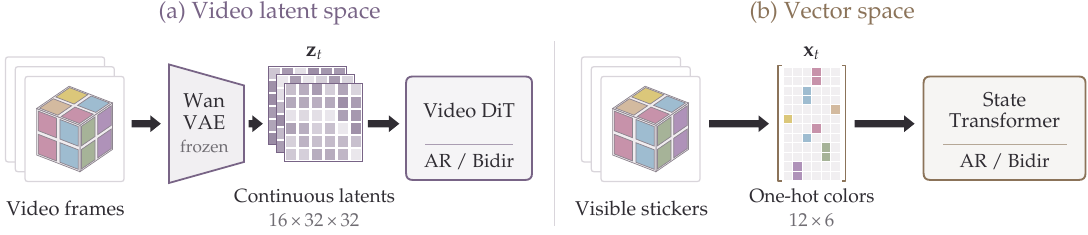}
    \caption{Video latent space and vector space. The action prompt is omitted. 
    (a) A frozen Wan VAE encodes video frames into continuous latents for the
    video DiT. (b) Visible sticker colors are represented directly as
    $12\times6$ one-hot arrays for the state Transformer.
    Both model families support autoregressive and bidirectional generation.}
    \label{fig:video-vector-spaces}
\end{figure}

To assess the difficulty of the underlying state-tracking task, we replace
video generation with direct prediction of sticker colors
(Figure~\ref{fig:video-vector-spaces}). We represent
the same 12 visible stickers as categorical variables and train Transformers
with approximately \emph{2.8M trainable parameters}, much smaller than the video DiT models. They receive the initial state
and action-prompt features from the shared frozen language encoder. Partial
observation is preserved: stickers that leave view must still be tracked
until they reappear. We evaluate sequences of \emph{20 actions}, not 9 in the video setting.
We train both models with batches of 256 episodes. The main training phase
ends when validation final-frame accuracy first exceeds 98\% or after 20,000
steps, whichever comes first. Each model then receives another 5,000 steps
at a lower learning rate. Both AR and bidirectional models learn accurate vector state prediction with much lower cost
(Table~\ref{tab:rubik-state-only}). This contrasts with the difficulty of maintaining the same
state information through video generation. Training details and learning
curves appear in Appendix~\ref{app:rubik-vector-space}.

\begin{table}[h]
    \centering
    \small
    \caption{Vector-space prediction on 10,000 held-out 20-action episodes.
    Compute estimates exclude the frozen language encoder.
    Final frame accuracy requires all 12 visible stickers to be correct after
    20 actions. Full-trajectory accuracy requires them to be correct
    after every action. AR uses its own predictions during rollout.}
    \label{tab:rubik-state-only}
    \begin{tabular}{lrrr}
        \toprule
        Generation & \shortstack{PF-days\\(approx.)}
        & \shortstack{Frame accuracy after\\\textbf{20 actions} (\%)}
        & \shortstack{Full-trajectory\\accuracy (\%)} \\
        \midrule
        AR    & $8\times10^{-3}$ & 94.01 & 92.81 \\
        Bidir & $4\times10^{-4}$ & 99.86 & 99.85 \\
        \bottomrule
    \end{tabular}
\end{table}

\paragraph{Full observation makes state reasoning easier}
\label{sec:full-observation}
To examine the difficulty introduced by partial observation, we compare
the usual three-face videos with synchronized complementary views that
expose all six faces. In the latter setting, every settled frame reveals
the full cube state, so hidden sticker colors do not have to be recovered
from earlier observations. This comparison uses a separate 270M-parameter
configuration with AR-$k=4$ (as this experiment was conducted in parallel with Stage A, before the final architecture was selected.) and bidirectional generation
(Appendix~\ref{app:rubik-full-observation-diagnostic}).
\begin{figure}[h]
    \centering
    \includegraphics[width=\linewidth]{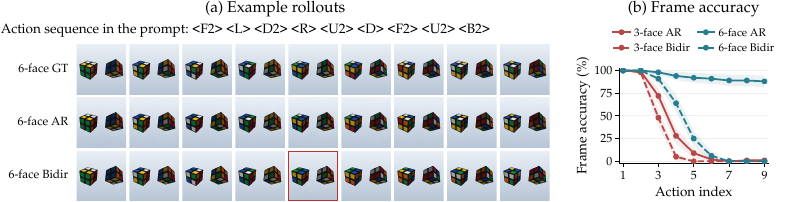}
    \caption{State prediction with three or six visible faces.
    (a) Ground truth and generated six-face videos after each action in the
    same 9-action sequence.
    Red boxes mark each model's first incorrect boundary frame.
    (b) Frame accuracy after each action, with 95\% normal confidence intervals
    over 100 paired episodes. Training exposure is matched between 3-face and 6-face. Frame accuracy requires all 12 visible stickers to be correct
    for three-face observation and all 24 for six-face observation.}
    \label{fig:rubik-observation-comparison}
    \label{fig:rubik-po-fo-qualitative}
    \label{fig:rubik-po-fo-short-horizon}
\end{figure}

Making all six faces visible substantially improves frame accuracy
(Figure~\ref{fig:rubik-observation-comparison}). AR maintains high accuracy
throughout the action sequence, while Bidir benefits mainly in the early
actions. The large improvement for AR suggests that tracking hidden
sticker states is a major source of difficulty: the model predicts sticker
transitions reliably when the full state remains visible. We next examine
whether more training data and larger models can improve prediction under
partial observation.

\subsection{Scaling Video Generator for Reasoning}
\paragraph{Additional Training Gives Diminishing Returns}

\begin{figure}[h]
    \centering
    \includegraphics[width=0.9\linewidth]{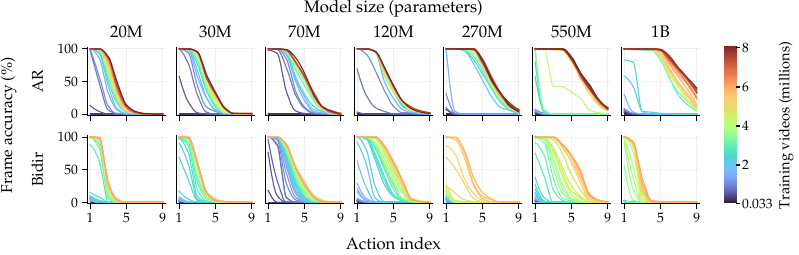}
    \caption{Frame accuracy after each action as training data increases,
    for AR (top) and Bidir (bottom).
    Colors indicate training videos on a shared scale. All selected
    checkpoints use the same 100 free-running episodes. Smooth lines
    connect the measured action accuracies.}
    \label{fig:action-frame-training-row}
\end{figure}

Training on more videos gives a model more experience with state
transitions. We ask whether this leads to sustained improvements in state
prediction when model size is held fixed.
Figure~\ref{fig:action-frame-training-row} shows that the gains diminish
as training progresses, with later checkpoints producing increasingly
similar accuracy profiles.
Similar diminishing returns have been observed when scaling training data
at a fixed model size, both in language modeling and in video
reasoning~\citep{kaplan2020scalinglaws,wang2026vbvr}.

\paragraph{Larger Models Improve Accuracy at a High Cost}

\label{sec:model-compute-scaling}

\begin{figure}[h]
    \centering
    \includegraphics[width=\linewidth]{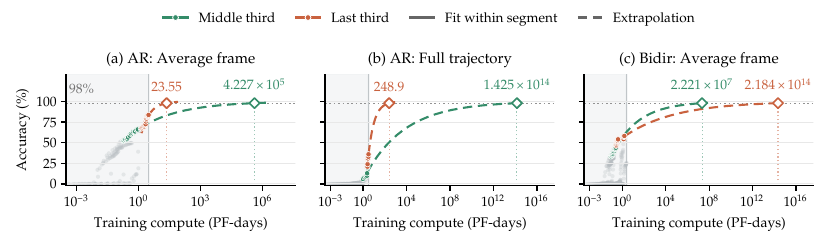}
    \caption{Illustrative compute extrapolations for (a) AR average frame accuracy,
    (b) AR full-trajectory accuracy, and (c) Bidir average frame accuracy.
    Gray points show evaluated checkpoints; colored points identify the
    middle and final thirds of each compute-ordered frontier.
    Solid lines show fits within each segment, and dashed lines show
    extrapolations. Diamonds mark 98\% accuracy, labeled in PF-days.
    Shading marks the observed compute range.
    Bidir full-trajectory accuracy is omitted because it is zero at every
    evaluated checkpoint.}
    \label{fig:state-compute-frontier}
\end{figure}

As gains from additional training diminish, larger models offer a way to
improve state prediction. We compare the best average frame and
full-trajectory accuracies observed within each compute budget, allowing
both model size and training duration to vary
(Figure~\ref{fig:state-compute-frontier}). Full-trajectory accuracy asks
whether a generated video remains correct through all nine actions,
rather than averaging correctness across individual frames.

Motivated by empirical power-law trends in classification error and
language-model compute scaling~\citep{hestness2017scaling,kaplan2020scalinglaws},
we fit power laws to prediction error along each observed frontier.
Following the window analysis of \citet{hoffmann2022training}, we compare
fits to the middle and final thirds of the frontier points to examine
how extrapolations depend on the fitting range
(Appendix~\ref{app:compute-training}).

Both fitting windows project substantially more compute to reach 98\%
accuracy, with complete AR trajectories demanding more than average frame
correctness. Their widely separated intersections make the numerical values
illustrative projections, not reliable training budgets. The common trend
is increasing compute demand as errors become rare, especially when every
state in a sequence must be correct.

\paragraph{MSE Does Not Reliably Compare Reasoning across Models}

\begin{figure}[h]
    \centering
    \includegraphics[width=\linewidth]{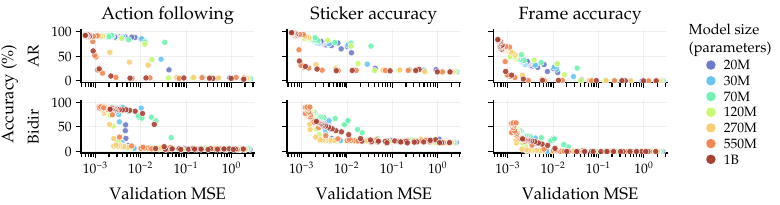}
    \caption{Validation MSE and reasoning accuracy across model sizes,
    for AR (top) and Bidir (bottom).
    Points pair validation MSE with free-running accuracy from the same EMA
    checkpoint. Colors identify model sizes.}
    \label{fig:mse-accuracy-reciprocal}
\end{figure}

Validation MSE follows an approximate power law as model size and training
data increase. It also tracks
broad learning progress within each model, with lower loss generally
accompanying higher accuracy.
Across models, however, MSE does not provide a common measure of capability.
Figure~\ref{fig:mse-accuracy-reciprocal} shows that similar losses can
correspond to substantially different action, sticker, and frame
accuracies. At 1.5M training videos, for example, the 1B AR model
has lower MSE than the 70M model, yet performs worse on all three metrics.
Lower MSE therefore does not reliably indicate better reasoning across
model sizes. Comparing models requires direct evaluation of action
execution and state correctness.

\subsection{Symbolic Guidance Improves Video Prediction}
\label{sec:state-guidance}

\begin{figure}[h]
    \centering
    \includegraphics[width=\linewidth]{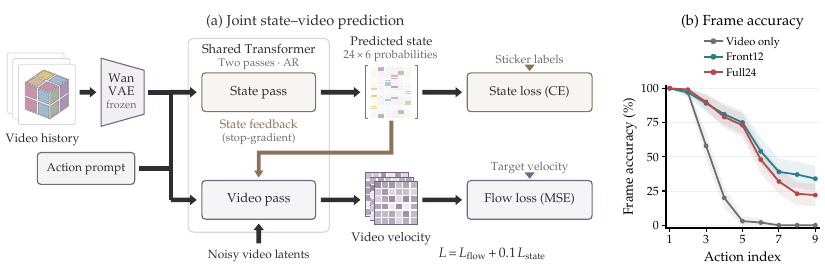}
    \caption{Explicit state guidance for 20M AR-$k=1$ models.
    (a) A shared Transformer predicts
    sticker-state distributions from the available video history and action
    prompt, then uses them to condition a separate video pass. State
    cross-entropy and video flow matching jointly train the model.
    (b) Frame accuracy after each action on the same 100 paired episodes
    after training on 3M videos. All curves use generated history.
    Shading shows 95\% confidence intervals.
    All models use EMA weights.}
    \label{fig:joint-state-video}
\end{figure}

The vector-space results suggest that learning an explicit symbolic
representation could help video generation. We test this idea by training
the video model to predict sticker states and using those predictions to
guide generation (Figure~\ref{fig:joint-state-video}(a)). The symbolic targets
specify the consequences of the prescribed actions, giving the model a
direct learning signal for state correctness alongside the video flow loss.

We compare the 20M AR-$k=1$ generator from the scaling study with two
state-guided variants, training each on the same 3M videos. Both variants
predict and supervise all 24 stickers. Front12 feeds back only the 12
visible-sticker distributions, while Full24 also includes the 12 hidden
stickers. We supervise categorical sticker colors and feed their predicted
probability distributions back to the video generator. During generation,
the model predicts these distributions from the action prompt and available
video history; simulator states are used only as training targets. Appendix~\ref{app:joint-state-video} gives the training
details.
State guidance substantially improves the accuracy of generated sticker
configurations (Figure~\ref{fig:joint-state-video}(b)). The 20M Front12 model
reaches 67.3\% average frame accuracy using 0.0445 PF-days, comparable to
67.1\% for the 550M video-only model at 1.33 PF-days. Front12 already achieves
the gain, so feeding back hidden-sticker predictions is not required when
all 24 stickers are supervised. Both variants sustain correct state
prediction further into the action sequence than the video-only model.
These results show that symbolic state supervision can improve the hidden state reasoning capability for video generation. Even though the handcrafted sticker representation may not generalize to other cases, it motivates learning
representations that capture symbolic state to guide video generation for reasoning capability

\section{Related Work}

\paragraph{Scaling prediction and generation.}
Language-model scaling laws describe prediction loss as a function of model
size, data, and compute~\citep{kaplan2020scalinglaws}.
Diffusion Transformers show improvements with increased model
compute~\citep{peebles2023scalable}. Subsequent studies fit explicit scaling
relationships for diffusion training loss and generation
metrics~\citep{liang2026diffusionscaling,chickering2026abra}, including
validation-loss relationships for video diffusion
Transformers~\citep{yin2025videoscaling}. We are inspired by those scaling experiments and evaluate if the video reasoning capability follows a similar trend.

\paragraph{Video evaluation and hidden-state tracking.}
VBench evaluates perceptual and temporal video quality~\citep{huang2023vbench};
VideoPhy and WorldModelBench assess physical commonsense and world-model
behavior~\citep{bansal2024videophy,li2025worldmodelbench}.
MBench and MemoBench emphasize memory and dynamically changing
environments~\citep{zhang2026mbench,chen2026memobench}.
\citet{shin2026video} investigate hidden-state tracking in an
action-conditioned Shell Game, particularly extrapolation beyond the training
horizon and mechanisms that support state updates.
VBVR and VBVR-Pro provide broad suites for training and evaluating visual
reasoning across tasks and models~\citep{wang2026vbvr,xu2026vbvrpro}.
We complement these suites with a controlled experiment that fixes the task
and prediction horizon, training models from scratch to study how scaling
affects hidden-state prediction rather than to benchmark frontier models.

\paragraph{State inference and semantic representations.}
Partially observable planning formalizes the need to use an
action--observation history when an observation does not identify the
underlying state~\citep{kaelbling1998planning}. Situation calculus and STRIPS
describe how actions transform that state~\citep{mccarthy1969philosophical,fikes1971strips,reiter2001knowledge}.
Predictive state representations express state through predictions of future
observations~\citep{littman2001predictive}. These perspectives motivate our
behavioral evaluation without requiring a symbolic representation.
REPA improves diffusion training by aligning denoiser representations with
clean-image features from a frozen pretrained visual
encoder~\citep{yu2025repa}. Our state-guidance experiment shares the motivation
of helping generation learn useful representations. We use symbolic state
supervision and condition video generation on predicted state distributions,
testing whether representations of action-dependent state changes improve
the correctness of generated outcomes.

\section{Conclusion}

We studied hidden-state reasoning through controlled Rubik's Cube video
prediction, scaling autoregressive and bidirectional models from 20M to 1B
parameters. Lower validation MSE does not reliably indicate more accurate
state prediction. Smaller autoregressive models perform better with limited
compute, while larger models reach higher accuracy after more training.
Small models readily learn state transitions in vector space, and revealing
the full state improves video prediction. Symbolic state supervision and
predicted-state feedback substantially improve accuracy at matched data
exposure. Our future work will seek generalizable and scalable symbolic
representations that capture how actions change state and guide reasoning
in video generation beyond handcrafted states for a single environment.

\subsubsection*{Acknowledgments}
LEK and WG have been supported in part by NSF 2336612 and Rice University Funds.

\clearpage
\bibliography{iclr2026_conference}
\bibliographystyle{iclr2026_conference}

\clearpage
\appendix
\section{Game Descriptions And Sanity Checks}

\label{app:benchmark-games}
\label{app:short-horizon-details}

\subsection{Rubik's Cube}

We study a solved \(2\!\times\!2\!\times\!2\) Rubik's Cube undergoing nine
prompted face turns. Each episode is rendered from a fixed three-quarter
camera as an 81-frame, \(256\!\times\!256\) video. At each time, the benchmark
camera exposes three faces (12 sticker positions) and occludes the other three.
Because the initial state, color convention, camera, and ordered action
sequence are fixed by the example, the state at every action boundary is
deterministic. The challenge is to infer the action-dependent sticker
configuration as stickers leave view and later reappear. Success requires history-dependent
prediction, but not an explicit 24-sticker internal representation.

Figure~\ref{fig:rubik-action-axes} illustrates the six face turns and their
rotation axes in the simulator coordinate system.

The first action samples uniformly from all six faces; subsequent actions
sample uniformly from the five faces other than the preceding face. For
each face, clockwise, counterclockwise, and half turns are equally likely.
This excludes immediate inverse cancellations. We apply no further
filtering by the resulting cube state. Training and evaluation use
disjoint episode seeds, and none of the 100 evaluation action sequences
appears in the 8M training pool.

We evaluate the initial frame and the first stationary frame after every move.
For each of the 12 visible stickers, the simulator supplies its expected color
and image region under the fixed camera. Within each region we take the median
RGB color and match it to the closest of the six cube colors by cosine
similarity. A sticker is marked unreadable, hence incorrect, if its RGB norm
is below 40 or its maximum color similarity is below 0.90. We report visible
sticker accuracy and exact-frame accuracy, which requires all 12 visible
stickers to be correct. The evaluator uses fixed simulator geometry rather
than a learned object detector or visual judge.

The frozen Wan VAE preserves every evaluated sticker configuration
(Table~\ref{tab:rubik-vae-state}). We apply the same color decoder to the
original videos and reconstructions of their cached ground-truth latents.

\begin{table}[h]
    \centering
    \small
    \caption{State accuracy (\%) on the same 100 paired evaluation videos
    before and after reconstruction with the frozen Wan VAE.
    All metrics use the nine
    post-action boundaries and exclude the initial frame.}
    \label{tab:rubik-vae-state}
    \begin{tabular}{lrrr}
        \toprule
        Video & Sticker & Frame & Full trajectory \\
        \midrule
        Original RGB & 100.00 & 100.00 & 100.00 \\
        Wan VAE reconstruction & 100.00 & 100.00 & 100.00 \\
        \bottomrule
    \end{tabular}

\end{table}

\subsection{Shared Language and Action Tokenization}

The vector and video generators use the same language-conditioning interface.
We extend a ModernBERT tokenizer with one token for each of the 18 legal moves:
six outer faces, each turned clockwise, counterclockwise, or \(180^\circ\).
For example, \texttt{<RUBIK\_R>} rotates the right face \(90^\circ\) clockwise
when viewed face-on, while \texttt{<RUBIK\_U\_PRIME>} rotates the upper face
counterclockwise. Scene instructions and the complete ordered action program
form a single prompt. We use an 8,192-token context window and reject, rather
than silently truncate, prompts that exceed it.

The shared language backbone is initialized from ModernBERT-base and frozen
before any generator is trained. For every appended action token, we
deterministically initialize its input embedding as the mean of the pretrained
embeddings of a short ordinary-language gloss of that action. The extended
embedding table is then frozen together with the rest of the language encoder;
neither the tokenizer nor the language backbone is optimized for this task.
The generator receives only the resulting language features and may learn a
model-specific projection into its hidden width; it receives no parallel
action-ID tensor, action-specific encoder, or structured simulator transition.
Thus the vector-only and video-only models share the information supplied by
language. Auxiliary semantic-state supervision, when used, is a separate
training intervention.

\subsection{Vanilla Video Generator}
\label{app:rubik-vanilla-generator}

The following historical 274M diagnostic is distinct from the Stage-A/B
scaling configurations. Its vanilla baseline is a latent rectified-flow
Transformer. The frozen Wan2.1 VAE maps each video to 21 latent frames with
16 channels and spatial
resolution \(32\times32\); the first latent frame is the visual condition and
the remaining 20 are prediction targets. The generator has 16 Transformer
blocks, width \(d_{\mathrm{model}}=1024\), 16 attention heads, and a SwiGLU
feedforward layer with ratio \(8/3\). Counting all trainable generator
parameters, the AR and bidirectional variants contain 273.9M and 273.8M
parameters, respectively. As in Table~\ref{tab:generator-parameter-compute},
the frozen VAE and ModernBERT encoder are excluded.

The bidirectional generator predicts all 20 target latent frames jointly. The
AR generator partitions them into five chronological chunks of four latent
frames, uses ground-truth prefixes during training, and feeds back its own
generated chunks at inference. We train both variants with AdamW, weight decay 0.05, gradient clipping
at 1.0, and EMA decay 0.9999. The learning rate warms up over the first 24k
episode presentations and then follows cosine decay to 10\% of its peak. The
peak learning rates are \(4.5\times10^{-4}\) for AR and
\(2.0\times10^{-4}\) for bidirectional generation. At evaluation, both use 16
midpoint sampling steps and the saved EMA generator parameters.
Figure~\ref{fig:rubik-po-fo-qualitative}(a) shows six-face rollouts for the
historical model at update 30,000. Exact-frame accuracy requires all visible
stickers to be correct.

\subsection{Joint State and Video Prediction}
\label{app:joint-state-video}

The state-guided models use the 20M AR-$k=1$ backbone in
Table~\ref{tab:scaling-runs}. Both variants have 19.81M trainable parameters.
Before generating each latent frame, ten learned queries predict the 24
sticker colors at the initial state and nine action boundaries from the
action prompt, initial video latent, and available video prefix.
The state and video passes share Transformer weights and use 3D RoPE.
A learned adapter maps predicted color probabilities to video conditioning,
with gradients stopped at the probabilities. Front12 masks the hidden
12 distributions before this adapter; Full24 retains all 24.

Each target latent uses the predicted state at the first action boundary
at or after its latent time, retaining the final boundary after the last
action. These targets describe settled action endpoints, including when
the video frame depicts a turn in progress. At inference, state predictions
are recomputed from the generated video prefix before each latent frame;
no simulator state is supplied as feedback. State cross-entropy supervises
all 24 stickers of the selected prediction at each of the 20 target latent
frames and is averaged over frames and stickers. The total loss is
$\mathcal{L}_{\mathrm{flow}}+0.1\mathcal{L}_{\mathrm{state}}$.

The video-only baseline and both state-guided models use the same 3M
training videos in the same order, shared backbone initialization, and
effective batch size 16, for 187,500 optimizer updates. All three use the
Stage-B optimizer and learning-rate
schedule in Appendix~\ref{app:video-dit-scaling}.
Including the state-query pass and feedback adapter, each guided model
uses an estimated 0.0445 PF-days for training on 3M videos, following the
FLOP convention in Appendix~\ref{app:compute-training}.
Figure~\ref{fig:joint-state-video}(b) evaluates the EMA checkpoints on the
same 100 paired episodes with matched sampling-noise identities and
16 midpoint steps per latent frame, excluding the initial frame.

\subsection{Observation Control: Three versus Six Faces}
\label{app:rubik-full-observation-diagnostic}

This diagnostic compares three-face observation with synchronized
complementary views exposing all six faces
(Figure~\ref{fig:rubik-po-fo-short-horizon}(b)). The comparison changes both
state visibility and the visual prediction target. We therefore report it
as an observation intervention, without attributing any difference solely
to memory or rendering capacity. The independent action-following check
below measures execution of the requested face turn separately.

Within each architecture, both observation settings use 30,000 updates and
the same global batch: 256 for AR and 48 for Bidir. These correspond to
7.68M and 1.44M video presentations, respectively, including repeated
training examples. The comparison matches exposure across views within
each architecture. AR and Bidir do not share a training-exposure budget. Therefore, we do not intend to show AR has a better performance than Bidir. 

Mean exact-frame accuracy rises from 34.67\% to 93.44\% for AR when moving
from three to six faces, while action-nine accuracy rises from 1\% to 88\%.
For Bidir, mean accuracy rises from 27.89\% to 42.89\%, while action-nine
accuracy remains 0\% in both settings. The mean-frame gains are 58.78
percentage points for AR (95\% paired-episode interval: [54.33, 62.67]) and
15.00 for Bidir ([12.56, 17.56]). These intervals use 2,000 bootstrap draws
(seed 20260915) over 100 matched initial states and action programs. Dataset
record identifiers differ across rendering configurations, so pairing is
verified from those physical episode contents. Exactness requires 12
stickers in the three-face view and 24 in the six-face view.

\subsection{Action-Following Sanity Check}
\label{app:rubik-action-following}

Poor sticker accuracy need not imply poor prompt following: a model may execute
the requested turn while applying it to an already incorrect cube state. We
therefore evaluate motion independently of sticker color. A frozen RAFT-Small
estimator~\citep{teed2020raft} extracts optical flow between two frames inside
each turn. We fit a linear 18-way probe for the moved face and turn type on
simulator videos whose six sticker colors are replaced by a new random
permutation of grayscale values in every episode. The probe therefore cannot
identify an action from the sticker palette. Throughout, \emph{action following}
requires both the face and turn type (clockwise \(90^\circ\), counterclockwise
\(90^\circ\), or \(180^\circ\)) to match the prompt. A segment without
detected motion is counted as incorrect.

Table~\ref{tab:rubik-action-following} evaluates the saved EMA parameters
of the historical 3M-video AR and bidirectional generators on 1,000 paired
held-out nine-action rollouts. The table reports both the full sequence and
action nine. Decoded ground-truth latents provide a separate VAE reconstruction
reference for the same frozen motion probe; that reference does not evaluate
generator weights.

At action nine, AR and Bidir correctly follow the requested motion in 80.6\% and 67.7\% of episodes, respectively, while their exact-frame accuracies on those same rollouts are 0.3\% and 0.0\%. Correct execution of the requested turn can therefore coexist with an incorrect sticker configuration.

\begin{table}[h]
    \centering
    \small
    \caption{\textbf{Rubik action following from grayscale motion.}
    The predicted face and turn type must both match the prompt. Values are
    percentages over 9,000 action segments, except the action-nine column, with
    1,000 segments.}
    \label{tab:rubik-action-following}
    \begin{tabular}{lrr}
\toprule
Source & Action following (all) & Action following (A9) \\
\midrule
VAE reconstruction & 92.6 & 92.1 \\
AR & 80.1 & 80.6 \\
Bidirectional & 69.6 & 67.7 \\
\bottomrule
\end{tabular}

\end{table}

\subsection{Vector-Space Formulation}
\label{app:rubik-vector-space}

\begin{figure}[h]
    \centering
    \includegraphics[width=0.5\linewidth]{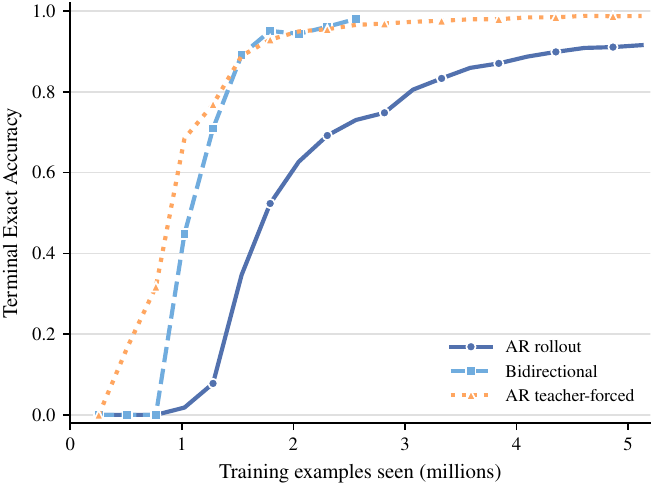}
    \caption{Vector-state learning curves for Rubik's Cube.}
    \label{fig:rubik-vector-learning-curve}
\end{figure}

To isolate sequence modeling from visual generation, we instantiate the same
task in a categorical vector space. The state \(\mathbf{x}_t\) contains one
six-way color variable for each sticker on the three faces visible from the
fixed camera. The \(2\times2\times2\) diagnostic therefore represents 12
visible stickers as a \(12\times6\) one-hot array. A sticker that moves out of
view is not represented again until it returns to a visible face, so success
still requires the model to maintain its intervening history.

The vector model receives only \(\mathbf{x}_0\) and the complete prompt from
the shared frozen language encoder, and predicts the state after every action.
It receives no rendered video, structured actions, simulator transition
operator, or intermediate oracle states. The bidirectional model predicts the
complete trajectory jointly. The AR model predicts one state at a time; it is
teacher-forced during training, while strict evaluation conditions each step
on its own preceding prediction.

We use 100,000 training episodes, 10,000 validation episodes, and 10,000
untouched test episodes, with 20 sampled actions per episode. Each prompt has
132 tokenizer tokens, including special tokens. The frozen
\texttt{ModernBERT-base} encoder conditions a six-layer
Transformer of width 192 with six attention heads and approximately 2.8M
trainable parameters. We optimize categorical cross-entropy with AdamW and
batches of 256 episodes. The main schedule warms up for 500 updates to
\(5\times10^{-4}\) and then decays by cosine toward \(5\times10^{-5}\).
Validation occurs every 1,000 updates. Main training stops at the first
validation with final-state success above 98\%, or at 20,000 updates, followed
by a fixed 5,000-update refinement from \(5\times10^{-5}\) to
\(3\times10^{-6}\). The bidirectional model uses 15,000 total updates; the AR
model uses 25,000.

For a rough compute estimate, we use a dense Transformer with one token per
sticker and a feedforward width of $4\times192$, including the prompt tokens.
We count a joint pass for Bidir and separate prefix passes for AR, and estimate
training FLOPs as three times the forward cost, excluding the frozen language
encoder.

For \(M\) test episodes, final-state success is
  $\mathrm{SR}_{\mathrm{Rubik}}
  =
  \frac{1}{M}\sum_{m=1}^{M}
  \mathbb{1}\!\left[
    \hat{\mathbf{x}}_{20}^{(m)}=\mathbf{x}_{20}^{(m)}
  \right]$.

Thus, all 12 visible sticker colors must be correct after move 20.
Table~\ref{tab:rubik-state-only} reports final-state and full-trajectory exact
accuracy on the untouched test set. In addition to the strict AR rollout
reported there, teacher-forced AR evaluation gives 99.17\% final-state success.

\clearpage

\section{Video DiT Architecture and Training Protocol}
\label{app:video-dit-scaling}
\label{app:proposed-scaling-law-study}

We scale only the trainable video diffusion Transformer (DiT); the frozen
Wan2.1 VAE and ModernBERT-base language encoder are excluded from all parameter
counts.  An 81-frame Rubik video is encoded into 21 latent times with 16
channels and spatial resolution $32\times32$.  A $1\times2\times2$ latent
patch therefore contains one latent time and produces 256 spatial tokens.  The
first latent time is the visual condition and the remaining 20 are prediction
targets, giving 5,120 predicted patch tokens per episode.  Autoregressive (AR)
models generate chunks of $k$ latent times, conditioned on their generated
history at inference; bidirectional models predict all 20 future latent times
jointly. Stage A evaluates $k\in\{1,4\}$; Stage B uses
AR with $k=1$.

Let $d_{\mathrm{model}}$ denote the Transformer width, $L$ the number of
blocks, $H$ the number of attention heads, $d_{\mathrm{head}}$ the width of
each head, and $d_{\mathrm{ff}}$ the SwiGLU hidden width.  Table~
\ref{tab:video-dit-architecture} defines the common architecture.  Only
$d_{\mathrm{model}}$ and $L$ vary with scale; Stage A determines their
allocation before Stage B varies model size and data.

\begin{table}[h]
  \centering
  \setlength{\tabcolsep}{4pt}
  \caption{Video DiT architecture used in the two-stage scaling study.}
  \label{tab:video-dit-architecture}
  \small
  \begin{tabularx}{\linewidth}{@{}lXl@{}}
    \toprule
    Component & Value or rule & Scaling treatment \\
    \midrule
    Transformer block & Self-attention, cross-attention, SwiGLU & Fixed \\
    Model width $d_{\mathrm{model}}$ & Tables~\ref{tab:stage-a-architecture} and~\ref{tab:scaling-runs}
      & Scaled \\
    Depth $L$ & Selected wide--shallow ladder & Scaled \\
    Position encoding & 3D RoPE & Fixed \\
    Head width $d_{\mathrm{head}}$ & $64$ & Fixed \\
    Attention heads $H$ & $d_{\mathrm{model}}/64$ & Derived \\
    FFN width $d_{\mathrm{ff}}$
      & $64\left\lceil(8d_{\mathrm{model}}/3)/64\right\rceil$ & Derived \\
    Latent representation & 16-channel frozen Wan2.1 VAE & Fixed \\
    Latent patch & $1\times2\times2$ & Fixed \\
    Text condition & Cached 768-dimensional ModernBERT features & Fixed \\
    Prediction factorization & Stage A: AR-$k\in\{1,4\}$ and bidirectional;
      Stage B: AR-$k=1$ & Fixed within each study \\
    \bottomrule
  \end{tabularx}
\end{table}

We use a two-stage design.  Stage A calibrates the depth--width allocation
before comparing parameter scales.  Stage B is the scaling-law experiment: it
holds the selected architecture family fixed and varies model size, data, and
training compute.  This calibration is specific to our long video-token
sequences; it prevents pathologies of an arbitrary depth--width rule from
being attributed to parameter scale.  Following the compute-oriented analysis
of DiT~\citep{peebles2023scalable}, the primary resource variable is training
FLOPs rather than parameter count alone, because AR history length and target
length also affect attention compute.

\paragraph{Stage A: architecture-family calibration.}
We compare the three parameter-matched configurations in
Table~\ref{tab:stage-a-architecture}.  Each is trained on the same
Rubik episode stream for up to 1M episode presentations with the same effective
batch size, optimizer, latent patch, and flow-time distribution.
Section~\ref{app:stage-a-results} reports the
completed nine-run AR-$k=1$/AR-$k=4$/bidirectional pilot at matched data
exposure, with per-configuration training-compute estimates in
Table~\ref{tab:stage-a-training-flops}. We select wide--shallow AR-$k=1$
on this data-matched evidence, not a claim of iso-FLOP optimality.
Throughput and peak memory are secondary systems measurements. This pilot's
1M cap does not set
the Stage-B training budget, and its runs do not contribute points to the
scaling-law fit.

\paragraph{Stage B: model--data scaling.}
We select the wide--shallow family from Stage A and retain seven approximately
geometric parameter targets. Table~\ref{tab:scaling-runs}
replaces the balanced fallback: widths increase and depths decrease while
parameter counts remain within 6\% of their original targets. Both width and
depth grow monotonically with scale (depth may stay constant), with at least
eight blocks. This preserves the design principle of allocating capacity
within a non-degenerate Transformer family, rather than imposing a constant
depth or claiming an optimal depth--width ratio. Parameter counts are exact
for the trainable generator, including latent, language, conditioning,
normalization, and output layers. The selection is supported by the
1M-video development comparison, not an established optimum at every
scale.

\begin{table}[h]
    \centering
    \caption{Stage B Video DiT architecture specifications.
    AR-$k=1$ and bidirectional models share the same architecture dimensions and parameter counts.}
    \label{tab:scaling-runs}
    \setlength{\tabcolsep}{4pt}
    \begin{tabular}{lrrrrr}
        \toprule
        Model name & Parameters (M) & $d_{\mathrm{model}}$ & Layers & Heads & $d_{\mathrm{ff}}$ \\
        \midrule
        20M  & 19.70 & 384  & 8  & 6  & 1024 \\ %
        30M  & 33.45 & 448  & 10 & 7  & 1216 \\ %
        70M  & 67.81 & 640  & 10 & 10 & 1728 \\ %
        120M & 115.8 & 768  & 12 & 12 & 2048 \\ %
        270M & 271.8 & 1088 & 14 & 17 & 2944 \\ %
        550M & 548.1 & 1408 & 17 & 22 & 3776 \\ %
        1B   & 938.4 & 1792 & 18 & 28 & 4800 \\ %
        \bottomrule
    \end{tabular}
\end{table}

Stage B varies the total trainable generator size $N$ and the number of
unique training episodes $D$, with training compute derived from the
attention-aware cost in Table~\ref{tab:generator-parameter-compute}.
All runs use the same episode prefix, seed 20260727, effective batch 16,
AdamW settings from Stage A, and a learning rate of $4\times10^{-4}$ after
16,384 videos of linear warmup. The learning rate then remains constant.
Packed chunk losses are averaged before one optimizer update; hardware
microbatching does not change the effective batch or the learning rate.

The scaling analysis contains 159 AR and 265 Bidir EMA checkpoints,
evaluated on the same 100 paired episodes. Each training video is consumed
once. The ground-truth-history diagnostic in
Appendix~\ref{app:stage-b-action-gt} uses a separate 140-checkpoint AR
snapshot, with model comparisons at a common exposure of 5M videos.

\subsection{Hyperparameter Selection}
\label{app:hyperparameter-selection}

We selected the effective batch size and peak learning rate using
matched-data pilot experiments on a 35.37M-parameter Video DiT with eight
blocks, width 512, and eight attention heads. Each run used the same 32,768
training videos, with linear warmup over 1,024 videos followed by cosine
decay to 10\% of the peak learning rate. Selection used final-checkpoint
validation flow MSE on 1,024 held-out videos, evaluated with non-EMA weights.

The batch-size search compares 16, 32, and 64 videos per update for
AR-$k=2$ at a peak learning rate of $2\times10^{-4}$. The learning-rate
search fixes the batch size at 16 and compares $10^{-4}$, $2\times10^{-4}$,
and $4\times10^{-4}$ for AR-$k\in\{1,2,4\}$ and bidirectional generation.
Batch size 16 and peak learning rate $4\times10^{-4}$ give the lowest
validation loss in their respective comparisons
(Table~\ref{tab:hyperparameter-search}); we use these settings in both
Stage A and Stage B.

\begin{table}[h]
    \centering
    \small
    \caption{Validation flow MSE in the learning-rate (left) and
    effective-batch-size (right) searches. Lower is better; best values
    within each comparison are bold.}
    \label{tab:hyperparameter-search}
    \begin{minipage}[t]{0.64\linewidth}
    \centering
    \setlength{\tabcolsep}{4pt}
    \begin{tabular}[t]{lrrr}
        \toprule
        Generation & \multicolumn{3}{c}{Peak learning rate} \\
        \cmidrule(lr){2-4}
        & $10^{-4}$ & $2\times10^{-4}$ & $4\times10^{-4}$ \\
        \midrule
        AR-$k=1$ & 0.1678 & 0.1217 & \textbf{0.0768} \\
        AR-$k=2$ & 0.1790 & 0.1337 & \textbf{0.0810} \\
        AR-$k=4$ & 0.1715 & 0.1287 & \textbf{0.0783} \\
        Bidir    & 0.1789 & 0.1274 & \textbf{0.0781} \\
        \bottomrule
    \end{tabular}
    \end{minipage}\hfill
    \begin{minipage}[t]{0.32\linewidth}
    \centering
    \setlength{\tabcolsep}{4pt}
    \begin{tabular}[t]{rr}
        \toprule
        Batch size & Validation MSE \\
        \midrule
        16 & \textbf{0.1337} \\
        32 & 0.1756 \\
        64 & 0.2409 \\
        \bottomrule
    \end{tabular}
    \end{minipage}
\end{table}

\clearpage
\section{Stage A: Architecture Calibration}
\label{app:stage-a-results}

\paragraph{Setup.}
Each of the three approximately 95M-parameter shapes in
Table~\ref{tab:stage-a-architecture} is trained with AR-$k=1$
(20 chunks), AR-$k=4$ (five chunks), or Bidir (20 future latent frames jointly).
All use 3D RoPE, frozen Wan2.1/ModernBERT representations, the full nine-action
prompt, and no vector-state input or auxiliary loss. Teacher-forced AR chunk
losses are computed in parallel and averaged before one update.

All runs use the same 1M distinct episodes, order, and training seed (20260727),
on one H200 per run. Batch size is 16 videos (81,920 target tokens), with no
gradient accumulation or activation checkpointing. AdamW uses betas
$(0.9,0.95)$, weight decay 0.05, and gradient clipping at 1.0. Learning rate
warms up over 16,384 videos to $4\times10^{-4}$, then decays by cosine to
$4\times10^{-5}$ at 1M. Flow times follow
$\operatorname{sigmoid}(\mathcal{N}(0,1))$. This cap applies only to Stage A.

\paragraph{Training compute.}
Table~\ref{tab:stage-a-training-flops} reports estimated training FLOPs for
each configuration: three times the forward matrix/convolution cost over
1M episodes, counting shared AR context projections once per episode.
Counts include trainable input/output projections and flow-time conditioning,
but exclude frozen encoders, optimizer/EMA updates, validation, and sampling.
The comparison matches data exposure, not training or sampling compute.

\begin{table}[htbp]
  \centering
  \small
  \caption{\textbf{Estimated Stage-A training compute}, in $10^{18}$ FLOPs,
  after 1M videos. One multiply--accumulate counts as two FLOPs; backward
  compute is approximated as twice forward compute. These are operation-count
  estimates, not hardware timings.}
  \label{tab:stage-a-training-flops}
  \begin{tabular}{lrrr}
\toprule
Shape & AR-$k=1$ & AR-$k=4$ & Bidir \\
\midrule
Deep--narrow & 5.015 & 5.281 & 6.343 \\
Balanced & 4.542 & 4.753 & 5.527 \\
Wide--shallow & 4.367 & 4.548 & 5.150 \\
\bottomrule
\end{tabular}

\end{table}

\paragraph{Evaluation.}
All 1M-video checkpoints (62,500 updates) use EMA weights, the same 100 held-out
development episodes and sampling-noise identities, and 16 midpoint steps per
AR chunk or joint Bidir video. AR uses generated history. Sticker accuracy
scores the 12 visible colors at settled post-action boundaries; exact-frame
accuracy requires all 12 correct. Action following requires the correct face
and turn type, measured by the independent grayscale-motion probe
(Section~\ref{app:rubik-action-following}). Absent motion counts as incorrect.
The same development episodes are used for the scaling comparisons.

\begin{table}[htbp]
  \centering
  \small
  \setlength{\tabcolsep}{4pt}
  \caption{\textbf{Stage A after 1M training videos.} Accuracies (\%) average
  actions 1--9 on the same 100 episodes, excluding the initial boundary.
  Flow MSE uses 1,024 validation episodes; its conditioning and fixed probes
  differ across prediction factorizations.}
  \label{tab:stage-a-1m-accuracy}
  \begin{tabular}{llrrrr}
\toprule
Shape & Family & MSE ($10^{-3}$) & Action & Sticker & Frame \\
\midrule
deep--narrow & AR-K1 & 2.233 & 17.22 & 25.37 & 3.67 \\
deep--narrow & AR-K4 & 2.562 & 6.33 & 22.20 & 0.78 \\
deep--narrow & BIDIR & 3.212 & 7.33 & 21.59 & 0.33 \\
balanced & AR-K1 & 1.883 & 72.44 & 52.00 & 24.33 \\
balanced & AR-K4 & 2.533 & 9.67 & 22.46 & 1.67 \\
balanced & BIDIR & 3.426 & 6.78 & 21.55 & 0.56 \\
wide--shallow & AR-K1 & 1.864 & 90.78 & 78.20 & 37.22 \\
wide--shallow & AR-K4 & 2.371 & 79.89 & 63.14 & 25.11 \\
wide--shallow & BIDIR & 3.184 & 8.33 & 21.67 & 0.78 \\
\bottomrule
\end{tabular}

\end{table}

\begin{figure}[htbp]
  \centering
  \includegraphics[width=\linewidth]{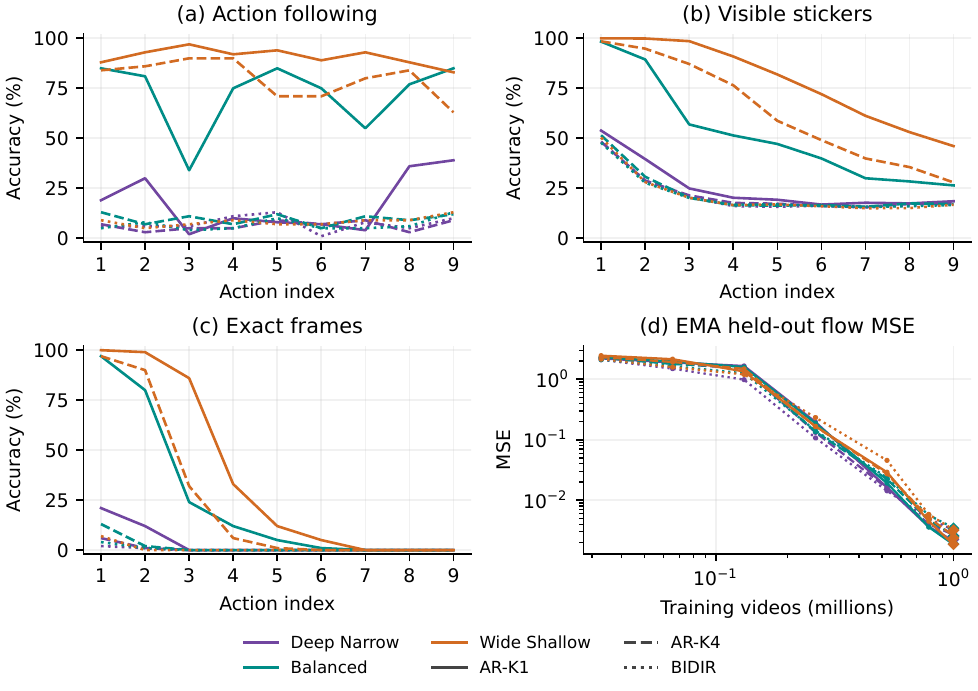}
  \caption{\textbf{Stage-A calibration across nine setups.}
  Purple, teal, and orange group deep--narrow,
  balanced, and wide--shallow; solid, dashed, and dotted lines distinguish
  AR-$k=1$, AR-$k=4$, and Bidir.
  \textbf{(a--c)} The 1M-video EMA checkpoints on 100 paired development episodes:
  action-following accuracy from the independent motion probe, visible-sticker
  accuracy, and exact-frame accuracy (all 12 stickers correct). The initial
  boundary is excluded.
  \textbf{(d)} EMA flow MSE on 256 fixed episodes at seven selected exposures:
  32,768, 65,536, 131,072, 262,144, 524,288, 786,432, and 1M videos.
  Diamond endpoint markers use 1,024 episodes. Initialization
  is omitted. AR loss uses ground-truth history; conditioning and noise/time
  probes differ across factorizations, so their losses are not directly
  comparable.}
  \label{fig:stage-a-1m-overview}
\end{figure}

\begin{figure}[htbp]
  \centering
  \includegraphics[width=\linewidth]{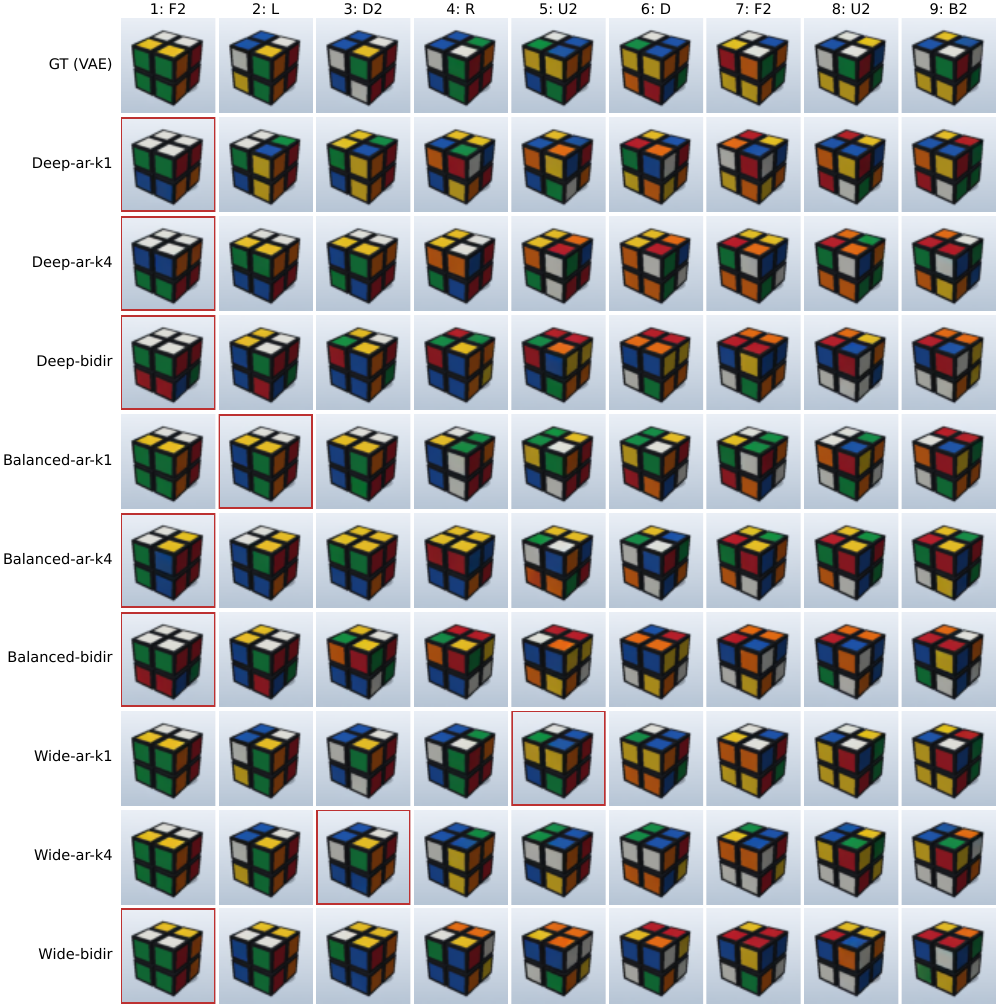}
  \caption{\textbf{Stage-A rollouts across all nine setups after 1M training videos.}
  VAE-reconstructed ground truth, followed by deep--narrow, balanced, and wide--shallow;
  each shape includes AR-$k=1$, AR-$k=4$, and Bidir.
  All use the first development episode (seed 81024000), the same prompt,
  and paired sampling-noise seeds, with 16 midpoint steps per AR chunk
  or entire Bidir video.
  The nine settled post-action boundaries are shown. Red boxes mark each model's
  first incorrect settled action boundary. The episode was not selected
  by performance. All panels use the saved EMA model parameters.}
  \label{fig:stage-a-rollout}
\end{figure}

\paragraph{Comparisons.}
Table~\ref{tab:stage-a-1m-accuracy} reports all nine EMA endpoints, while
Figure~\ref{fig:stage-a-1m-overview} separates action following, sticker
accuracy, and exact-frame accuracy along the action sequence. Wide--shallow
AR-$k=1$ has the highest point estimates for all three metrics among these
nine configurations, with 90.78\% action-following and 37.22\% exact-frame
accuracy. These EMA endpoints support carrying wide--shallow AR-$k=1$ into
Stage B, where we keep the architecture family fixed while varying model size.
These comparisons describe the observed training budget and do not establish
an asymptotic convergence floor or reasoning ceiling.
Figure~\ref{fig:stage-a-rollout} shows all nine setups on the first fixed
development episode, alongside the aggregate measurements.

\clearpage
\section{Full Scaling Results}
\label{app:stage-b-results}

The main scaling analysis uses 159 AR and 265 Bidir EMA checkpoints,
each evaluated on the same 100 paired episodes. Action, sticker, and frame
accuracies average the nine post-action boundaries. Full-trajectory accuracy requires all
nine post-action frames in an episode to be correct.
The ground-truth-history diagnostic below uses a separate 140-checkpoint
AR snapshot and 16 midpoint steps per AR-$k=1$ chunk.

\subsection{Action-Aligned Ground-Truth History}
\label{app:stage-b-action-gt}

All 140 diagnostic checkpoints have paired GT-history evaluations. For each
separately tested action, the GT prefix ends at the last causal VAE latent
strictly before the action begins. The generator predicts the remaining
0--3 bridge frames and the entire action without further GT refresh; known
GT past also supplies decoder context. Each pair shares model weights,
episode identities, sampling noise, and the 16-step midpoint sampler.
The nine GT windows are initialized independently. Free-running scores
instead come from a continuous rollout. The protocol preserves the action
durations, latent timing, and AR chunk size.

\begin{table}[h]
\centering
\small
\begin{tabular}{lrrrr}
\toprule
Model & Mean frame: Free & GT prefix & A9 frame: Free & GT prefix\\
\midrule
XS & 33.89 & 38.67 & 0.00 & 5.00\\
S & 42.22 & 47.89 & 0.00 & 9.00\\
M & 52.56 & 60.00 & 0.00 & 22.00\\
B & 48.11 & 55.00 & 1.00 & 9.00\\
L & 63.56 & 71.67 & 3.00 & 38.00\\
XL & 64.56 & 73.33 & 5.00 & 45.00\\
XXL & 71.00 & 78.67 & 10.00 & 45.00\\
\bottomrule
\end{tabular}

\caption{Matched free-running and action-aligned GT-history comparisons at
5M videos. All values are percentages on the same 100 episodes.}
\label{tab:gt-prefix-5m}
\end{table}

\clearpage
\section{Generator Compute and Training Details}
\label{app:compute-training}

We scale a latent rectified-flow Transformer by the number of blocks $L$ and
the hidden width $d_{\mathrm{model}}$. Each block contains visual attention
over the target and permitted visual context, language cross-attention, and a SwiGLU
layer of width
$d_{\mathrm{ff}}=64\lceil 8d_{\mathrm{model}}/(3\cdot64)\rceil$.  The
concatenated attention width equals
$d_{\mathrm{model}}$; each head has width 64, hence
$n_{\mathrm{heads}}=d_{\mathrm{model}}/64$.  The block-only parameter count $N_{\mathrm{blocks}}$ excludes projections.
The empirical scaling analysis uses the full trainable count $N$. The block
count is approximately:
\begin{equation}
    N_{\mathrm{blocks}} \simeq L\left(8d_{\mathrm{model}}^2
    +3d_{\mathrm{model}}d_{\mathrm{ff}}\right),
    \label{eq:generator-parameters}
\end{equation}
where the two attention modules contribute $8d_{\mathrm{model}}^2$ and the
SwiGLU contributes $3d_{\mathrm{model}}d_{\mathrm{ff}}$ per block.

For compute, each cached video latent frame has $d_z$ channels and spatial size
$H_z\times W_z$.  Dividing it into $p\times p$ patches gives
$S=(H_z/p)(W_z/p)$ visual tokens per frame.  A chunk of $k$ target frames
therefore contains $n_{\mathrm{tgt}}=kS$ target tokens.  Its context contains
$n_{\mathrm{text}}$ cached language tokens, $n_{\mathrm{anchor}}=S$ tokens from
the initial latent frame, and $n_{\mathrm{hist}}$ tokens from preceding latent
frames, so
$n_{\mathrm{ctx}}=n_{\mathrm{text}}+n_{\mathrm{anchor}}+n_{\mathrm{hist}}$.
Each cached language token has feature width $d_f$.  We count one
multiply--accumulate as two FLOPs and omit lower-order operations, following
\citet{kaplan2020scalinglaws}.

\begin{table}[h]
    \centering
    \small
    \setlength{\tabcolsep}{4pt}
    \caption{\textbf{Generator parameters and forward compute.}
    FLOPs are per target visual patch token; biases, normalization, and
    nonlinearities are omitted.}
    \label{tab:generator-parameter-compute}
    \resizebox{\linewidth}{!}{%
    \begin{tabular}{@{}lcc@{}}
        \toprule
        Operation & Parameters & Forward FLOPs per target token \\
        \midrule
        Language projection
            & $d_f d_{\mathrm{model}}$ & $2n_{\mathrm{text}}d_f d_{\mathrm{model}}/n_{\mathrm{tgt}}$ \\
        Latent input/output projections
            & $3d_zp^2d_{\mathrm{model}}$ & $2d_zp^2d_{\mathrm{model}}\left(2+\frac{n_{\mathrm{anchor}}+n_{\mathrm{hist}}}{n_{\mathrm{tgt}}}\right)$ \\
        Target visual Q/K/V and output
            & $4Ld_{\mathrm{model}}^2$ & $8Ld_{\mathrm{model}}^2$ \\
        Visual-context K/V (shared weights)
            & --- & $4L\frac{n_{\mathrm{anchor}}+n_{\mathrm{hist}}}{n_{\mathrm{tgt}}}d_{\mathrm{model}}^2$ \\
        Visual attention products
            & --- & $4L(n_{\mathrm{tgt}}+n_{\mathrm{anchor}}+n_{\mathrm{hist}})d_{\mathrm{model}}$ \\
        Language cross-attention projections
            & $4Ld_{\mathrm{model}}^2$ & $4Ld_{\mathrm{model}}^2\left(1+\frac{n_{\mathrm{text}}}{n_{\mathrm{tgt}}}\right)$ \\
        Language attention products
            & --- & $4Ln_{\mathrm{text}}d_{\mathrm{model}}$ \\
        SwiGLU feedforward
            & $3Ld_{\mathrm{model}}d_{\mathrm{ff}}$ & $6Ld_{\mathrm{model}}d_{\mathrm{ff}}$ \\
        Flow-time MLP
            & $3d_{\mathrm{model}}^2$ & $6d_{\mathrm{model}}^2/n_{\mathrm{tgt}}$ \\
        \midrule
        Total (Transformer blocks)
            & $N_{\mathrm{blocks}}\simeq L(8d_{\mathrm{model}}^2+3d_{\mathrm{model}}d_{\mathrm{ff}})$
            & $L\!\left[12d_{\mathrm{model}}^2+4\frac{n_{\mathrm{ctx}}}{n_{\mathrm{tgt}}}d_{\mathrm{model}}^2+6d_{\mathrm{model}}d_{\mathrm{ff}}+4d_{\mathrm{model}}(n_{\mathrm{tgt}}+n_{\mathrm{ctx}})\right]$ \\
        \bottomrule
    \end{tabular}}
\end{table}

The frozen video autoencoder and language encoder are excluded from $N$.
For packed AR, shared language, latent-context, and context-K/V projections
are counted once per episode, over all materialized context tokens; attention
products use only the current noisy chunk and its permitted clean prefix.
For Bidir, all future tokens attend to one another and to the initial frame.
We estimate training FLOPs as three times this forward cost (one forward and
approximately two backward passes). Optimizer/EMA updates, validation, and
sampling are excluded. In AR, causal masks prevent access to future latent frames.

For Figures~\ref{fig:rubik-scaling-overview}(c)
and~\ref{fig:state-compute-frontier}, we construct a separate observed
frontier for each metric and generation family. We sort checkpoints by
training compute and retain the initial baseline and each subsequent
record improvement in accuracy. We then divide the complete frontier
into three equal groups of points and fit the middle and final thirds
separately. Each segment contains 16 points for AR average frame accuracy,
6 for AR full-trajectory accuracy, and 14 for Bidir average frame accuracy.
This window comparison is motivated by the frontier curvature examined
by \citet{hoffmann2022training}.

For accuracy $a$, we fit $1-a=AC^{-\alpha}$ by ordinary least squares in
log space. Here $C$ is training compute in PF-days, with one PF-day equal
to $8.64\times10^{19}$ FLOPs. The fitted curve intersects target
accuracy $q$ at $C_q=(A/(1-q))^{1/\alpha}$. These intersections illustrate
the implications of extending the fitted trends; they are not validated
predictions of the compute needed at the target accuracy.
Table~\ref{tab:compute-segment-fits} reports the fitted coefficients and
98\% intersections. Average frame accuracy averages the nine post-action
correctness indicators, whereas full-trajectory accuracy requires all
nine to be correct within the same episode. Bidir has no successful full
trajectories in these evaluations, so we do not extrapolate that metric.

\begin{table}[h]
    \centering
    \small
    \caption{Power-law fits to the middle and final thirds of each observed
    compute frontier. $C_{98}$ is the 98\% intersection of each fitted
    curve, in PF-days, illustrating its sensitivity to the fitting window.}
    \label{tab:compute-segment-fits}
    \begin{tabular}{@{}llrrr@{}}
\toprule
Metric & Segment & $A$ & $\alpha$ & $C_{98}$ \\
\midrule
AR frame & Middle third & 0.336 & 0.2178 & $4.227\times10^{5}$ \\
AR frame & Final third & 0.5607 & 1.055 & $23.55$ \\
AR trajectory & Middle third & 0.9501 & 0.1185 & $1.425\times10^{14}$ \\
AR trajectory & Final third & 1.59 & 0.7931 & $248.9$ \\
Bidir frame & Middle third & 0.4366 & 0.1823 & $2.221\times10^{7}$ \\
Bidir frame & Final third & 0.4424 & 0.09379 & $2.184\times10^{14}$ \\
\bottomrule
\end{tabular}

\end{table}

We convert compute to H100 GPU-hours as $24C/(0.5\times0.989)$, using the
50\% model FLOPs utilization assumption of \citet{sardana2024beyond}
and a dense BF16 peak of 989 TFLOP/s.\footnote{NVIDIA's throughput specifications:
\url{https://github.com/NVIDIA/exemplar-performance\#peak-theoretical-throughput}.}

\subsection{Training Procedure}
\label{app:training-procedure}

For minibatch $\mathcal{B}=\{(z_0,z,y)\}$, let $z_0$ be the initial-frame
latent, $z=(z_1,\ldots,z_{n_{\mathrm{frame}}})$ the future latents, and $y$ the
language features.  For clean target $\tilde z$, noise $\epsilon\sim\mathcal{N}(0,I)$,
and $t\sim\operatorname{sigmoid}(\mathcal{N}(0,1))$, rectified-flow training uses
$x_t=(1-t)\epsilon+t\tilde z$ and target velocity $\tilde z-\epsilon$.  Algorithm~\ref{alg:generator-training}
shows one optimizer update; $\operatorname{MSE}$ averages over the minibatch and
all latent elements. After each optimizer update, we update a float32
exponential moving average (EMA) of every trainable generator parameter:
\[
    \bar\theta_t=0.9999\,\bar\theta_{t-1}+0.0001\,\theta_t.
\]
The EMA starts from the initialized generator parameters and is restored
alongside the optimizer when training resumes. Video-generator evaluations
use the EMA saved in each selected checkpoint, including flow-MSE probes,
free rollouts, GT-prefix interventions, and qualitative samples. The frozen
language/video encoders are unchanged. Parameter EMA is distinct from any
smoothing of plotted optimization losses.

\begin{algorithm}[h]
\begin{minipage}[t]{0.47\linewidth}
\centering
\textbf{Bidirectional}
\begin{algorithmic}[1]
\Require $\mathcal{B}$, generator $G_\theta$
\State Sample $t,\epsilon$; form $x_t$ for $z$
\State $\hat v \gets G_\theta(x_t,t\mid z_0,y)$
\State $\mathcal{L}\gets\operatorname{MSE}(\hat v,z-\epsilon)$
\State Update $\theta$ using $\nabla_\theta\mathcal{L}$
\State $\bar\theta\gets0.9999\bar\theta+0.0001\theta$
\end{algorithmic}
\end{minipage}\hfill
\begin{minipage}[t]{0.51\linewidth}
\centering
\textbf{Autoregressive}
\begin{algorithmic}[1]
\Require $\mathcal{B}$, $G_\theta$, chunk size $k$
    \State Partition $z$ into $n_{\mathrm{chunk}}=n_{\mathrm{frame}}/k$ chunks
    \ForAll{$i\in\{1,\ldots,n_{\mathrm{chunk}}\}$ \textbf{in parallel}}
        \State Sample $t_i,\epsilon_i$; form $x_{t_i}^{(i)}$ for $z^{(i)}$
        \State $c_i\gets(z_0,y,z^{(<i)})$ \Comment{ground-truth prefix}
        \State $\hat v^{(i)}\gets G_\theta(x_{t_i}^{(i)},t_i\mid c_i)$
        \State $\mathcal{L}_i\gets\operatorname{MSE}(\hat v^{(i)},z^{(i)}-\epsilon_i)$
    \EndFor
    \State $\mathcal{L}\gets n_{\mathrm{chunk}}^{-1}\sum_{i=1}^{n_{\mathrm{chunk}}}\mathcal{L}_i$
    \State Update $\theta$ once using $\nabla_\theta\mathcal{L}$
    \State $\bar\theta\gets0.9999\bar\theta+0.0001\theta$
\end{algorithmic}
\end{minipage}
\caption{Bidirectional and autoregressive generator training.}
\label{alg:generator-training}
\end{algorithm}

\end{document}